\documentclass[11pt]{article}
\usepackage[final]{acl}
\usepackage{times}
\usepackage{latexsym}
\usepackage[T1]{fontenc}
\usepackage[utf8]{inputenc}
\usepackage{microtype}
\usepackage{inconsolata}
\usepackage{graphicx}
\usepackage{amsmath,amssymb,amsthm}
\usepackage{mathtools}
\usepackage{algorithm}
\usepackage{algpseudocode}
\usepackage{booktabs}
\usepackage{hyperref}
\usepackage{cleveref}
\usepackage{bm}
\usepackage{bbm}
\usepackage{xcolor}
\usepackage{thmtools}
\usepackage{multirow}

\title{SVD-Attention: Efficiently Analyzing Attention Blocks in Transformers}

\author{ \textbf{Debopam Sanyal}$^{1}$ \quad \textbf{Hongjie
  Chen}$^{2}$ \quad \textbf{Riya Elizabeth Kuriyan}$^{1}$ \quad
  \textbf{Arth Tak}$^{1}$ \\
  \textbf{Divya Kothandaraman}$^{2}$ \quad \textbf{Akshay Mehra}$^{2}$
  \quad \textbf{Joshua Kimball}$^{2}$ \quad \textbf{Alexey
  Tumanov}$^{1}$ \\
  $^{1}$Georgia Institute of Technology \quad $^{2}$Dolby Laboratories
  }

\begin{document}
\maketitle

\begin{abstract}
Pretrained transformers spend a large share of their inference compute
inside self-attention, yet the per-head query-key interaction is
structurally low-rank by construction. We ask a diagnostic question:
across layers and architectures, which heads actually use the full
rank budget on real data, and which do not? We propose SVD-Attention,
a post-hoc method that decomposes the per-head interaction matrix
$M_h$ and scores every singular mode by both its spectral magnitude
and its activation variance across tokens on a small calibration set.
The score keeps the modes the head can amplify and that real data
actually activates, and discards the rest; the resulting factorization
is exact for the retained modes and needs no finetuning. Across seven
pretrained models and five task types, the diagnostic finds a
task-dependent pattern: redundancy sits in the early to middle layers
for sentiment and image classification and for generation, and in the
late layers for dense prediction and extractive question answering. It
is also predictive: truncating the layers it flags costs at most
$1.75$ F1, while truncating the same number of layers elsewhere in the
same model costs up to $50$ F1. SVD-Attention is an analysis tool
rather than a compression method: acting on it reduces inference FLOPs
by $0.8$-$1.5\%$ on its own and by $8$-$11\%$ when composed with
weight pruning, with $0$-$4\%$ change in task quality.
\end{abstract}

\section{Introduction}
\label{sec:intro}

Multi-head self-attention is the central building block of modern
transformers, and a long line of work has sought to reduce its cost by
changing the attention pattern~\cite{beltagy2020longformer,
liu2021swin}, approximating the softmax
kernel~\cite{wang2020linformer, choromanski2020rethinking,
meng2025polaformer, fan2025breaking}, or removing redundancy from
pretrained projection weights after training~\cite{yuan2023asvd,
wang2025svd, sanyal2026klas}. We follow this third direction. Existing
post-hoc methods decompose $W_{q,h}$ and $W_{k,h}$ independently,
which leaves their product, the object the softmax actually consumes,
unconstrained. We instead decompose the per-head interaction matrix
$M_h = W_{q,h} W_{k,h}^{\top}$ directly. $M_h$ has rank at most $d_h$,
so the head has algebraic room for many directions but uses far fewer
in practice: a singular direction only affects the head's output if
real input tokens vary along it, and otherwise contributes a constant
offset that softmax absorbs. We outline our method in
Fig.~\ref{fig:overview}.

We therefore score each mode of $M_h$ by the product of its singular
value and the geometric mean of its query- and key-side activation
variances on a small calibration set. Keeping the smallest set of
modes that captures a $(1-\epsilon)$ fraction of that importance gives
an exact rank-$r_h$ factorization with no finetuning. We use this
diagnostic to ask three questions. First, does adding data to the
score change which modes look important compared to ranking by
singular value alone (\textbf{RQ1})? Second, where in a pretrained
transformer does the diagnostic find the most redundancy, and does the
pattern depend on the task (\textbf{RQ2})? Third, is the diagnostic
predictive: when we remove the modes it flags, does performance hold
up where the score is low and drop where it is high (\textbf{RQ3})?

We run the diagnostic on seven pretrained transformers spanning
language understanding (DistilBERT), extractive question answering
(RoBERTa-base, MiniLM), generation (Llama-2-7B), vision classification
(ViT, Swin), and dense prediction (DPT-Large). Redundancy sits in the
early to middle layers for sentiment and image classification and for
generation, and in the late layers for dense prediction and extractive
question answering. Acting on the diagnostic reduces inference FLOPs
by $0.8$-$1.5\%$ on its own; composing it with weight
pruning~\cite{sun2024wanda} raises the reduction to $8$-$11\%$, with
task quality within $0$-$4\%$ of baseline and no gradient updates. In
the flagged layers more than half of the average head's rank budget
can be discarded for $0.16$ F1 (RoBERTa-base on SQuAD 2.0), and the
same diagnostic predicts where truncation instead fails.

\section{Related Work}
\label{sec:related}

\textbf{Efficient attention.}
The quadratic cost of softmax attention has motivated linear and
sparse approximations~\cite{wang2020linformer,
choromanski2020rethinking, liu2021swin}, and recent work continues to
close the gap with softmax~\cite{meng2025polaformer, fan2025breaking}.
These methods change the attention computation itself. Our work is
orthogonal: we keep softmax attention intact and study redundancy in
the pretrained projection weights.

\textbf{Token-level compression.}
A complementary direction reduces the number of tokens rather than the
per-token cost: Token Merging groups similar tokens at each layer and
increases throughput with negligible accuracy
drop~\cite{bolya2022token}. Token merging shrinks the sequence length
while our method shrinks the per-head channel dimension $d_h$, so the
two axes are complementary.

\textbf{Post-training SVD of weights.}
Closest to our setting, ASVD~\cite{yuan2023asvd} compresses LLM
projections by reweighting columns according to activation magnitudes
before SVD, and SVD-LLM~\cite{wang2025svd} adds a truncation-aware
whitening step that links singular value loss to output loss. Both
compress each weight matrix in isolation and are not specific to
attention. We differ in three ways. First, our unit of decomposition
is the per-head interaction matrix $M_h$, the object the softmax
consumes; compressing $W_{q,h}$ and $W_{k,h}$ separately leaves their
product unconstrained. Second, our score uses activation variance in
the singular bases, which captures discriminability rather than
magnitude: a mode contributing a near-constant offset to every token
is washed out by softmax even when its activations are large. Third,
we use the construction as a diagnostic for where redundancy lives
across layers and tasks, not only as a compression algorithm.

\textbf{Magnitude-based pruning.}
WANDA~\cite{sun2024wanda} prunes individual weights by the product of
weight magnitude and input activation norm, with no retraining. WANDA
and our method are both training-free and calibration-based but target
different units: WANDA removes individual entries within each weight
matrix, while we remove whole singular modes of $M_h$. We compose with
WANDA as a stronger baseline in \S\ref{sec:results}.

\textbf{Attention rank and head pruning.}
\citet{elhage2021mathematical} analyze the combined QK matrix as a
circuit object, the same object we decompose, and
\citet{bhojanapalli2020low} study the per-head rank budget as an
expressivity limit; our finding is the complement, that real data
often does not use a budget that is algebraically sufficient.
\citet{michel2019sixteen} and \citet{voita2019analyzing} prune whole
heads, a coarser unit than ours: a head that is indispensable as a
whole can still spend most of its rank budget on modes that no token
varies along. DRONE~\cite{chen2021drone} and
FWSVD~\cite{hsu2022language} are data-aware low-rank compressors, but
like ASVD and SVD-LLM their objective is reconstruction error, which
ranks a spectrally large but data-inactive mode highly.

\section{SVD-Attention Method}
\label{sec:method}

\begin{figure}[!t]
\centering
\includegraphics[width=0.99\columnwidth]{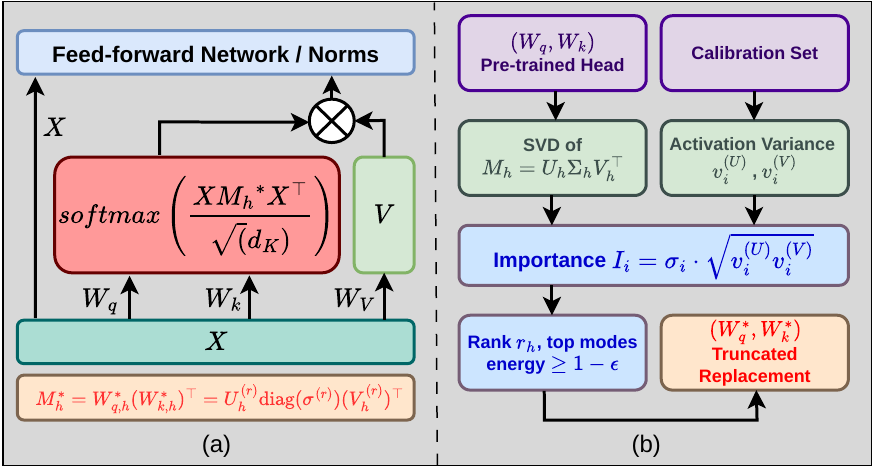}
\caption{Conceptual representation of our proposed SVD-Attention is shown in (a). The SVD-Attention method is outlined in (b).}
\label{fig:overview}
\end{figure}

Fig.~\ref{fig:overview} outlines the method;
Section~\ref{app:algorithm} gives the complete algorithm.

\subsection{Setup}
Let $X \in \mathbb{R}^{N \times d}$ be the input to a self-attention
layer with $H$ heads, head dimension $d_h = d / H$, and per-head
projections $W_{q,h}, W_{k,h} \in \mathbb{R}^{d \times d_h}$. The
attention logits inside head $h$ depend on $X$ only through the
interaction matrix
\begin{equation}
M_h \;=\; W_{q,h} W_{k,h}^{\top} \;\in\; \mathbb{R}^{d \times d},
\label{eq:interaction}
\end{equation}
since $Q_h K_h^{\top} = X M_h X^{\top}$. By construction
$\mathrm{rank}(M_h) \le d_h$, so $M_h$ lives in a subspace of
dimension at most $d_h$ inside a $d \times d$ matrix. We ask whether
real input data uses this rank budget fully, or whether some modes are
unused in practice and can be removed.

\subsection{Decomposition}
We decompose $M_h$ directly with the singular value decomposition:
\begin{equation}
M_h \;=\; U_h \, \Sigma_h \, V_h^{\top},
\label{eq:svd}
\end{equation}
where $U_h, V_h \in \mathbb{R}^{d \times d}$ are orthogonal and
$\Sigma_h = \mathrm{diag}(\sigma_1, \ldots, \sigma_d)$ with $\sigma_1
\ge \cdots \ge \sigma_d \ge 0$. Because $\mathrm{rank}(M_h) \le d_h$,
at most $d_h$ singular values are nonzero. The columns of $U_h$ form a
basis for queries and those of $V_h$ a basis for keys; the two bases
are in general distinct.

\subsection{Data-Aware Importance}

The singular value $\sigma_i$ measures how much mode $i$ can amplify a
query-key interaction, but not whether real tokens differ along that
mode. A direction along which all tokens project to the same value
adds a near-constant offset to every attention logit and is washed out
by the softmax, so it contributes nothing to the head's output
regardless of its singular value. We capture this with a data-aware
variance term.

Given a small calibration set $\mathcal{C} =
\{X^{(c)}\}_{c=1}^{|\mathcal{C}|}$ of layer inputs, we project each
example into the two bases and measure how much tokens vary along each
singular direction:
\begin{equation}
v_i^{(U)} = \frac{1}{|\mathcal{C}|} \sum_{c} \mathrm{Var}_t\big[(X^{(c)} U_h)_{t,i}\big],
\label{eq:varU}
\end{equation}
and analogously $v_i^{(V)}$ using $V_h$, with the variance taken
across tokens $t \in \{1, \ldots, N\}$ and averaged over calibration
examples. We combine the weight-side capacity and the data-side
activation into a single importance score:
\begin{equation}
\mathcal{I}_i \;=\; \sigma_i \cdot \sqrt{v_i^{(U)} \, v_i^{(V)}}.
\label{eq:importance}
\end{equation}
The geometric mean is the right combination because it goes to zero
whenever either side is zero: if queries vary along mode $i$ but all
keys project to the same value, the dot product $q_{s,i} \, k_{t,i}$
is constant across key tokens and the mode produces no
differentiation~\cite{yuan2023asvd}. The arithmetic mean would give
such a mode misleadingly moderate importance.

\subsection{Adaptive Rank Selection and Replacement}
\label{sec:rank}

We sort modes by $\mathcal{I}_i$ in descending order and pick the
smallest $r_h$ such that the top modes capture at least a $(1 -
\epsilon)$ fraction of the linear importance:
\begin{equation}
r_h \;=\; \min \Bigg\{ r \,:\, \frac{\sum_{i=1}^{r} \mathcal{I}_{(i)}}{\sum_{i=1}^{d} \mathcal{I}_{(i)}} \;\ge\; 1 - \epsilon \Bigg\},
\label{eq:rank}
\end{equation}
where $\mathcal{I}_{(i)}$ denotes the $i$-th largest score. Each head
chooses its own rank, so heads that use less of the rank budget are
compressed more aggressively. Let $U_h^{(r)}, V_h^{(r)} \in
\mathbb{R}^{d \times r_h}$ collect the columns of $U_h$ and $V_h$ at
the top-$r_h$ modes and let $\sigma^{(r)} = (\sigma_{\pi(1)}, \ldots,
\sigma_{\pi(r_h)})$ collect their singular values. We replace the
per-head projections by
\begin{equation}
\resizebox{0.89\columnwidth}{!}{$
\displaystyle
W^{*}_{q,h} = U_h^{(r)} \, \mathrm{diag}(\sqrt{\sigma^{(r)}}), \quad
W^{*}_{k,h} = V_h^{(r)} \, \mathrm{diag}(\sqrt{\sigma^{(r)}}).
$}
\label{eq:replace}
\end{equation}
This factorization is exact for the retained modes: $M^{*}_h =
W^{*}_{q,h} (W^{*}_{k,h})^{\top} = U_h^{(r)} \,
\mathrm{diag}(\sigma^{(r)}) \, (V_h^{(r)})^{\top}$, the truncated SVD
of $M_h$ at rank $r_h$. The only approximation error comes from the
discarded modes, and is bounded (\S\ref{app:proof}).

\subsection{Inference Cost}
\label{sec:cost}

The replacement weights have shape $d \times r_h$ instead of $d \times
d_h$, so each compressed head reduces both the projection FLOPs and
the attention-logit FLOPs by a factor of $r_h / d_h$. The
decomposition is performed once offline, and the reported reduction
assumes the discarded query and key dimensions are not computed. For
models using rotary position embeddings (RoPE), such as Llama-2-7B, we
apply the same post-hoc truncation and evaluate it empirically
(\S\ref{app:rope}). We report analytic FLOPs throughout, as the
post-training compression literature we build on does; realizing them
as wall-clock speedup requires an attention kernel operating at the
reduced dimension $r_h$, since padding the discarded dimensions back
erases the saving.

\subsection{Truncation Error Bound}
\label{app:proof}

We bound the spectral-norm error introduced by truncating $M_h$ to
rank $r_h$ using our data-aware ordering.

\paragraph{Setup.}
Let $M_h = U_h \Sigma_h V_h^{\top}$ be the SVD of the per-head
interaction matrix, with singular values $\sigma_1 \ge \cdots \ge
\sigma_d \ge 0$. Let $\pi$ be the permutation that sorts modes by
$\mathcal{I}_i = \sigma_i \sqrt{v_i^{(U)} v_i^{(V)}}$ in descending
order, and let $\mathcal{S} = \{\pi(1), \ldots, \pi(r_h)\}$ be the
indices of the retained modes. Define
\begin{equation}
\widetilde M_h \;=\; \sum_{i \in \mathcal{S}} \sigma_i \, u_i v_i^{\top}.
\end{equation}
This is exactly the rank-$r_h$ matrix produced by
Algorithm~\ref{alg:svdattn}.

\paragraph{Proposition.}
The spectral-norm truncation error satisfies
\begin{equation}
\big\| M_h - \widetilde M_h \big\|_2 \;=\; \max_{i \notin \mathcal{S}} \sigma_i.
\label{eq:spec_bound}
\end{equation}

\paragraph{Proof.}
By construction $M_h - \widetilde M_h = \sum_{i \notin \mathcal{S}}
\sigma_i \, u_i v_i^{\top}$. The vectors $\{u_i\}_{i \notin
\mathcal{S}}$ are mutually orthonormal because they are columns of
$U_h$; the same holds for $\{v_i\}_{i \notin \mathcal{S}}$. Hence $M_h
- \widetilde M_h$ has SVD $\sum_{i \notin \mathcal{S}} \sigma_i \, u_i
v_i^{\top}$ already in canonical form, and its spectral norm equals
its largest singular value, $\max_{i \notin \mathcal{S}} \sigma_i$.
$\square$

\paragraph{Comparison with Eckart--Young.}
The Eckart--Young theorem~\cite{eckart1936approximation} gives
\begin{equation}
\min_{\mathrm{rank}(B) \le r_h} \| M_h - B \|_2 \;=\; \sigma_{r_h + 1}.
\end{equation}
The minimum is attained by retaining the top-$r_h$ singular values in
spectral order. When the data-aware ordering agrees with the spectral
ordering, our truncation is optimal in the spectral norm. When it
disagrees, the gap is the difference between our retained set and the
top-$r_h$ spectral set:
\begin{equation}
\big\| M_h - \widetilde M_h \big\|_2 \;-\; \sigma_{r_h+1} \;=\; \big(\max_{i \notin \mathcal{S}} \sigma_i\big) - \sigma_{r_h+1}.
\label{eq:gap}
\end{equation}
The right-hand side is non-negative and is zero when no high-spectral
mode is discarded in favor of a lower-spectral one.

\paragraph{Why we accept this gap.}
We trade spectral-norm optimality for a property the spectral ranking
lacks: modes that are spectrally large but data-inactive contribute a
near-constant offset to the attention logits. Concretely, if all
tokens project to nearly the same value $\bar c_i$ along mode $i$,
then mode $i$'s contribution to the logit $q_s^{\top} k_t$ is
approximately $\sigma_i \bar c_i^2$ for every $(s, t)$ pair. A
constant added to every entry of a softmax row leaves the output
distribution unchanged, so retaining such a mode wastes rank budget
without changing attention. The variance term in $\mathcal{I}_i$ is
the data-side analogue of $\sigma_i$ and quantifies precisely how much
each mode contributes \emph{beyond} a constant offset. The gap in
Eq.~\ref{eq:gap} therefore upper-bounds a quantity we explicitly do
not want to minimize.

\paragraph{Frobenius bound.}
The same argument applies to the Frobenius norm:
\begin{equation}
\big\| M_h - \widetilde M_h \big\|_F^2 \;=\; \sum_{i \notin \mathcal{S}} \sigma_i^2.
\end{equation}
The Eckart-Young Frobenius lower bound is $\sum_{i > r_h} \sigma_i^2$,
and the data-aware Frobenius error exceeds this by exactly the
spectral mass we sacrifice to keep data-active modes.

\subsection{SVD-Attention with RoPE}
\label{app:rope}

Llama-2-7B applies rotary position embeddings
(RoPE)~\cite{su2024roformer} to the queries and keys after the linear
projections. For positions $s$ and $t$, the RoPE-augmented queries and
keys are
\begin{equation}
\tilde q_s = R_s q_s, \quad \tilde k_t = R_t k_t,
\end{equation}
where $q_s = (X W_{q,h})_s$, $k_t = (X W_{k,h})_t$, and $R_s,R_t \in
\mathbb{R}^{d_h \times d_h}$ are orthogonal, position-dependent
rotations. The resulting attention score is
\begin{equation}
\tilde q_s^{\top} \tilde k_t
=
q_s^{\top} R_s^{\top} R_t k_t
=
(XW_{q,h})_s^{\top} R_s^{\top} R_t (XW_{k,h})_t
\label{eq:rope_score}
\end{equation}
Unlike standard attention, the effective query-key interaction is
therefore not fully described by a single position-independent matrix
$M_h$: the bilinear form depends on relative position through
$R_s^{\top}R_t$.

We nonetheless compute the SVD of $M_h$ and use it as a diagnostic
proxy for the query-key interaction in RoPE models. After selecting
the retained modes, we replace the query and key projections with the
truncated factors
\begin{equation}
\resizebox{0.89\columnwidth}{!}{$
W^{*}_{q,h} = U_h^{(r)} \mathrm{diag}(\sqrt{\sigma^{(r)}}),
\quad
W^{*}_{k,h} = V_h^{(r)} \mathrm{diag}(\sqrt{\sigma^{(r)}}),
$}
\end{equation}
or, when an implementation expects the original head dimension $d_h$,
their zero-padded versions. The standard RoPE operation is then
applied to the resulting queries and keys without changing the RoPE
module itself, inducing the score
\begin{equation}
(\tilde q^{*}_s)^{\top}\tilde k^{*}_t
=
(XW^{*}_{q,h})_s^{\top} R_s^{\top} R_t (XW^{*}_{k,h})_t ,
\end{equation}
which has the same RoPE form as the original model. This is a
practical post-hoc approximation, not an exact low-rank factorization
of the full position-dependent interaction; the exact factorization
statement in \S\ref{sec:rank} applies to the position-independent
matrix $M_h$. The three arguments below say why the approximation is
nonetheless well founded.

\paragraph{Mode selection is position-independent.}
Writing $R_s^{\top} R_t = R_{D}$ with $D = t-s$, the RoPE logit is
$x_s^{\top} M^{(D)} x_t$ with $M^{(D)} = W_{q,h} R_D W_{k,h}^{\top}$,
so the effective interaction is a relative-position family and $M_h$
is its $D=0$ member. But $R_D$ is orthogonal and acts inside the head
space, after the projection. Hence for every $D$ the column space of
$M^{(D)}$ lies in that of $W_{q,h}$, its row space in that of
$W_{k,h}$, and $\mathrm{rank}(M^{(D)}) \le d_h$. RoPE re-weights
directions the head already reads; it adds no rank and cannot make the
head read an input direction that $W_{q,h}$ or $W_{k,h}$ does not
span. Which input directions a head reads is what our diagnostic
measures, and the activation variances behind it are computed on the
layer input $X$, which RoPE never touches, so the selection criterion
is position-independent by construction.

\paragraph{The truncation error is bounded uniformly in position.}
Let $q^{*}_s$ and $k^{*}_t$ be the queries and keys produced by the
truncated projections. The logit error splits into two cross terms,
each a product of a truncation error on one side and a rotated but not
rescaled vector on the other. Because $R_D$ preserves length,
Cauchy-Schwarz gives
\begin{equation}
\begin{aligned}
&\bigl| \tilde q_s^{\top} \tilde k_t
  - (\tilde q^{*}_s)^{\top} \tilde k^{*}_t \bigr| \\
&\qquad \le
  \| q_s - q^{*}_s \| \, \| k_t \|
  + \| q^{*}_s \| \, \| k_t - k^{*}_t \|,
\end{aligned}
\label{eq:rope_bound}
\end{equation}
which contains no $D$. A rotation changes the orientation of the error
but not its magnitude, so the failure mode a nontrivial interaction
would produce, an error that is small at short range and grows with
distance, is precisely what the isometry precludes. This relies on the
same $R_D$ acting on the exact and on the truncated queries and keys,
which holds because RoPE is applied in the original head coordinates.

\paragraph{Changes for a discarded mode.}
What RoPE does change is the fate of a discarded mode. If every token
projects to nearly the same value along mode $i$, its contribution to
the logit is approximately $\sigma_i \, \bar c^{(U)}_i \bar c^{(V)}_i
\, (u_i^{\top} R_D v_i)$. This varies with $D$ but is identical for
every token pair at that offset. Without RoPE it is a constant that
softmax absorbs exactly; with RoPE it becomes a fixed positional bias
that varies with distance but carries no information about which
tokens sit at those positions. Discarding such a mode therefore
removes at most a positional bias, never content-dependent signal.
This is weaker than exact absorption, and we state it as such.

\paragraph{Empirical check.}
in Tab.~\ref{tab:qa_flagged}, truncating the flagged Llama-2-7B layers
$\{8,\ldots,11\}$ keeps WikiText-2 perplexity within $0.11$ of
baseline ($5.58$ against $5.47$) and MMLU within $0.65$ points
($44.65$ against $45.30$), while truncating the non-flagged layers
$\{4,\ldots,7\}$ costs $2.64$ perplexity ($8.11$) and $4.45$ MMLU
($40.85$). A proxy breaking down under rotation would degrade both;
instead the diagnostic separates them cleanly.

\paragraph{Grouped-query attention.}
Llama-2-7B uses standard multi-head attention. For grouped-query
attention (GQA), where multiple query heads share a key/value head, a
RoPE-aware extension would also need to account for the shared key
projection across the group. A fully RoPE-aware variant would further
need to account for the rotary coordinate structure explicitly, for
example by selecting complete rotary coordinate pairs or by
decomposing the relative-position-dependent family rather than the
single matrix $M_h$. We leave both to future work.

\subsection{Algorithm}
\label{app:algorithm}

Algorithm~\ref{alg:svdattn} summarizes the full \textsc{SVD-Attention}
procedure for one attention layer and one head. The procedure is
applied independently across heads and across the layers selected by
the diagnostic. The calibration pass that produces
$\{X^{(c)}\}_{c=1}^{|\mathcal{C}|}$ is a single forward pass per
example with hooks attached at the input of each attention layer, run
once before any decomposition.

\begin{algorithm}[h]
\caption{\textsc{SVD-Attention} for one head}
\label{alg:svdattn}
\begin{algorithmic}[1]
\Require Pretrained projections $W_{q,h}, W_{k,h} \in \mathbb{R}^{d
\times d_h}$; calibration inputs $\{X^{(c)}\}_{c=1}^{|\mathcal{C}|}$
at this layer; threshold $\epsilon$ \Ensure Replacement projections
$W^{*}_{q,h}, W^{*}_{k,h} \in \mathbb{R}^{d \times r_h}$ \State $M_h
\gets W_{q,h} W_{k,h}^{\top}$ \Comment{interaction matrix} \State
$U_h, \Sigma_h, V_h^{\top} \gets \mathrm{SVD}(M_h)$ \Comment{$\Sigma_h
= \mathrm{diag}(\sigma_1, \ldots, \sigma_d)$} \For{$c = 1, \ldots,
|\mathcal{C}|$} \State $\hat X^{(c)}_U \gets X^{(c)} U_h$ \State $\hat
X^{(c)}_V \gets X^{(c)} V_h$ \EndFor \For{$i = 1, \ldots, d$} \State
$v_i^{(U)} \gets \tfrac{1}{|\mathcal{C}|}\sum_{c} \mathrm{Var}_t\big[
\hat X^{(c)}_{U, :, i}\big]$ \State $v_i^{(V)} \gets
\tfrac{1}{|\mathcal{C}|}\sum_{c} \mathrm{Var}_t\big[ \hat X^{(c)}_{V,
:, i}\big]$ \State $\mathcal{I}_i \gets \sigma_i \cdot \sqrt{v_i^{(U)}
\, v_i^{(V)}}$ \EndFor \State Sort modes by $\mathcal{I}_i$ in
descending order, giving permutation $\pi$ \State $r_h \gets
\min\big\{ r : \sum_{i \le r}\mathcal{I}_{(i)} \big/ \sum_i
\mathcal{I}_{(i)} \ge 1-\epsilon \big\}$ \State $U_h^{(r)} \gets
[U_{h, :, \pi(1)} \cdots U_{h, :, \pi(r_h)}]$ \State $V_h^{(r)} \gets
[V_{h, :, \pi(1)} \cdots V_{h, :, \pi(r_h)}]$ \State $\sigma^{(r)}
\gets (\sigma_{\pi(1)}, \ldots, \sigma_{\pi(r_h)})$ \State
$W^{*}_{q,h} \gets U_h^{(r)} \, \mathrm{diag}(\sqrt{\sigma^{(r)}})$
\State $W^{*}_{k,h} \gets V_h^{(r)} \,
\mathrm{diag}(\sqrt{\sigma^{(r)}})$ \State \Return $W^{*}_{q,h},
W^{*}_{k,h}$
\end{algorithmic}
\end{algorithm}

\section{Experimental Setup}
\label{sec:setup}

We evaluate on seven pretrained transformers across five task types:
DistilBERT~\cite{sanh2019distilbert} on SST-5~\cite{socher2013sst} for
sentiment classification (Accuracy);
RoBERTa-base~\cite{liu2019roberta} and MiniLM~\cite{wang2020minilm} on
SQuAD 2.0~\cite{rajpurkar2018squad2} and
NewsQA~\cite{trischler2017newsqa} for extractive question answering
(F1); Llama-2-7B~\cite{touvron2023llama2} on
TruthfulQA~\cite{lin2022truthfulqa} (ROUGE-L),
MMLU~\cite{hendrycks2021mmlu} (5-shot Accuracy) and
WikiText-2~\cite{merity2017pointer} (perplexity) for generation;
Swin-B~\cite{liu2021swin} and ViT-B/16~\cite{dosovitskiy2021vit} on
ImageNet-1K~\cite{deng2009imagenet} for image classification at
$224\times 224$ (Top-1); and DPT-Large~\cite{ranftl2021dpt} on
ADE20K~\cite{zhou2017ade20k} for semantic segmentation (mIoU). All
models are public checkpoints; we apply no finetuning or architectural
changes.

\textbf{Calibration.}
We draw 20000, 2000, 200, and 50 examples from the training
split/benchmarks of ImageNet-1K, SST-5, TruthfulQA, and ADE20K
respectively, and compute the per-mode variances in one forward pass
per example with hooks at each attention layer. For the extractive QA
models we calibrate on $1{,}303$ SQuAD 2.0 examples ($1\%$ of train)
unless stated otherwise; \S\ref{app:robustness} shows the
layer-level answer is unchanged from $100$ to $19{,}548$ examples.

\textbf{Knobs and baselines.}
The diagnostic exposes an energy threshold $\epsilon$ and a set of
target layers; we sweep $\epsilon \in \{0.1,\ldots,0.6\}$ and report
representative configurations (layer-wise sweep in
\S\ref{app:sweep}). We compare against the unmodified model,
against WANDA~\cite{sun2024wanda} at sparsity $0.1$, and against our
method composed with WANDA on top, using the same calibration set
throughout.

\textbf{Metrics.}
We report task quality and the change in inference FLOPs relative to
the unmodified model, computed from the per-head retained rank $r_h$
and post-WANDA sparsity.

\section{Results}
\label{sec:results}

\subsection{Answering the Research Questions}
\label{sec:rq}

\begin{table}[t]
\centering
\small
\resizebox{\columnwidth}{!}{%
\begin{tabular}{lccc}
\toprule
\multicolumn{4}{l}{\textit{(a) ViT-B, ImageNet-1K, layers $\{6,7,8\}$
(Top-1 \%)}} \\
\midrule
Method & $\epsilon=0.1$ & $\epsilon=0.2$ & $\epsilon=0.3$ \\
\midrule
Singular value only                    & $77.84$ & $77.13$ & $76.12$
\\
\textbf{SVD-Attention (data-aware)}    & $\mathbf{77.94}$ &
$\mathbf{77.62}$ & $\mathbf{76.86}$ \\
\midrule
Gap                                    & $+0.10$ & $+0.49$ & $+0.74$
\\
\bottomrule
\end{tabular}
}

\resizebox{\columnwidth}{!}{%
\begin{tabular}{llcccc}
\toprule
\multicolumn{6}{l}{\textit{(b) RoBERTa-base at $\epsilon=0.3$ (F1,
mean retained rank of 64)}} \\
\midrule
Dataset & Layers & $r_h$ (sv) & F1 (sv) & $r_h$ (ours) & F1 (ours) \\
\midrule
SQuAD 2.0 & $\{8,\ldots,11\}$ & $38.6$ & $63.08$ & $29.0$ &
$\mathbf{82.85}$ \\
NewsQA    & $\{8,\ldots,11\}$ & $38.6$ & $59.33$ & $27.0$ &
$\mathbf{69.06}$ \\
NewsQA    & $\{4,\ldots,7\}$  & $38.5$ & $20.73$ & $28.8$ &
$\mathbf{59.77}$ \\
\bottomrule
\end{tabular}
}
\caption{Ablation of the ranking rule, holding the unit of decomposition and the layer selection fixed. (a) On ViT-B the data-aware score wins at every $\epsilon$ and the gap widens as more modes are discarded. (b) On RoBERTa-base the singular-value-only rule keeps roughly ten more modes per head and still loses up to $39$ F1.}
\label{tab:svd_only_vs_data_aware}
\end{table}

\textbf{RQ1: Does data add information beyond singular value?}
To isolate the variance term, we compare our ranking against ranking
by $\sigma_i$ alone at three values of $\epsilon$
(Tab.~\ref{tab:svd_only_vs_data_aware}a). The data-aware score wins at
every $\epsilon$, and the gap widens monotonically: $+0.10$ Top-1 at
$\epsilon=0.1$, $+0.49$ at $\epsilon=0.2$, and $+0.74$ at
$\epsilon=0.3$. This is the expected signature of a useful data-aware
term: at small $\epsilon$ few modes are discarded and the two rankings
nearly agree, but as more are removed the singular-value-only ranking
starts discarding modes that real data uses. The effect is far larger
on RoBERTa-base at $\epsilon=0.3$
(Tab.~\ref{tab:svd_only_vs_data_aware}b), where that rule keeps
roughly ten more modes per head and still loses up to $39$ F1: on
NewsQA layers $\{4,\ldots,7\}$ it scores $20.73$ against our $59.77$.
A mode can therefore be spectrally large and contribute only a
near-constant offset, which is precisely the mode a
reconstruction-based objective would retain and ours discards.

\begin{table}[!t]
\centering
\small
\resizebox{\columnwidth}{!}{%
\begin{tabular}{lccrr}
\toprule
Model & Layers & $\epsilon$ & Metric & $\Delta$ FLOPs \\
\midrule
\multicolumn{5}{l}{\textit{DistilBERT on SST-5 (\textbf{Accuracy
\%})}} \\
\quad Baseline                & ---       & ---  & $49.95$ & ---
\\
\quad WANDA-only              & ---       & ---  & $49.82$ & -$9.65\%$
\\
\quad Ours                    & $\{0,1,2\}$ & 0.1  & $50.95$ &
-$0.83\%$   \\
\quad Ours + WANDA            & $\{0,1,2\}$ & 0.1  & $51.09$ &
-$10.39\%$  \\
\midrule
\multicolumn{5}{l}{\textit{RoBERTa-base on NewsQA (\textbf{F1})}} \\
\quad Baseline                & ---       & ---  & $70.81$ & ---
\\
\quad Ours                    & $\{8,\ldots,11\}$ & 0.1 & $70.74$ &
-$1.47\%$  \\
\quad Ours                    & $\{8,\ldots,11\}$ & 0.3 & $69.06$ &
-$3.63\%$  \\
\quad Ours + WANDA            & $\{8,\ldots,11\}$ & 0.1 & $70.61$ &
-$10.33\%$ \\
\quad Ours + WANDA            & $\{8,\ldots,11\}$ & 0.3 & $69.04$ &
-$12.32\%$ \\
\midrule
\multicolumn{5}{l}{\textit{Llama-2-7B on TruthfulQA
(\textbf{ROUGE-L})}} \\
\quad Baseline                & ---       & ---  & $0.5362$ & ---
\\
\quad WANDA-only              & ---       & ---  & $0.5359$      &
-$9.55\%$      \\
\quad Ours                    & $\{8,\ldots,11\}$ & 0.1 & $0.4952$ &
-$1.13\%$      \\
\quad Ours + WANDA            & $\{8,\ldots,11\}$ & 0.1 & $0.4948$
& -$10.57\%$      \\
\midrule
\multicolumn{5}{l}{\textit{ViT-B on ImageNet-1K (\textbf{Top-1 \%})}}
\\
\quad Baseline                & ---       & ---  & $81.08$ & ---
\\
\quad WANDA-only              & ---       & ---  & $81.02$ & -$9.45\%$
\\
\quad Ours                    & $\{6,7,8\}$ & 0.1 & $77.94$ &
-$0.76\%$   \\
\quad Ours + WANDA            & $\{6,7,8\}$ & 0.1 & $77.94$ &
-$10.13\%$ \\
\midrule
\multicolumn{5}{l}{\textit{Swin-B on ImageNet-1K (\textbf{Top-1 \%})}}
\\
\quad Baseline                & ---       & ---  & $84.71$ & ---
\\
\quad WANDA-only              & ---       & ---  & $84.67$ & -$9.49\%$
\\
\quad Ours                    & $\{4,\ldots,7\}$ & 0.1 & $84.68$ &
-$1.29\%$   \\
\quad Ours + WANDA            & $\{4,\ldots,7\}$ & 0.1 & $84.67$ &
-$10.65\%$ \\
\midrule
\multicolumn{5}{l}{\textit{DPT-Large on ADE20K (\textbf{mIoU})}} \\
\quad Baseline                & ---       & ---  & $45.99$ & ---
\\
\quad WANDA-only              & ---       & ---  & $46.02$ & -$6.98\%$
\\
\quad Ours                    & $\{16,\ldots,19\}$ & 0.1 & $46.02$ &
-$1.54\%$ \\
\quad Ours + WANDA            & $\{16,\ldots,19\}$ & 0.1 & $46.02$ &
-$8.40\%$ \\
\bottomrule
\end{tabular}
} \caption{Main results across seven pretrained transformers and five
task types; settings whose metrics do not fit this format are reported
in Tab.~\ref{tab:qa_flagged}. \textit{Ours} applies SVD-Attention
(data-aware SVD truncation); \textit{Ours + WANDA} applies WANDA
pruning at sparsity $0.1$ on top. Metric measures task quality in
units written in \textbf{bold}. $\Delta$ FLOPs is the \% reduction in
inference FLOPs relative to baseline.}
\label{tab:main}
\end{table}

\begin{table}[t]
\centering
\small
\resizebox{\columnwidth}{!}{%
\begin{tabular}{lccc}
\toprule
\multicolumn{4}{l}{\textit{(a) Extractive QA at $\epsilon=0.3$ (F1)}}
\\
\midrule
Model / dataset & Baseline & Flagged & Non-flagged \\
\midrule
RoBERTa / SQuAD 2.0 & $83.01$ & $\mathbf{82.85}$ \tiny$\{8..11\}$ &
$52.23$ \tiny$\{4..7\}$ \\
RoBERTa / NewsQA    & $70.81$ & $\mathbf{69.06}$ \tiny$\{8..11\}$ &
$58.85$ \tiny$\{0..3\}$ \\
MiniLM / SQuAD 2.0  & $77.77$ & $\mathbf{77.12}$ \tiny$\{9..11\}$ &
$69.98$ \tiny$\{6..8\}$ \\
MiniLM / NewsQA     & $56.03$ & $\mathbf{55.58}$ \tiny$\{9..11\}$ &
$\phantom{0}5.94$ \tiny$\{3..5\}$ \\
\bottomrule
\end{tabular}
}

\resizebox{\columnwidth}{!}{%
\begin{tabular}{lccc}
\toprule
\multicolumn{4}{l}{\textit{(b) Llama-2-7B at $\epsilon=0.1$}} \\
\midrule
Metric & Baseline & Flagged $\{8..11\}$ & Non-flagged $\{4..7\}$ \\
\midrule
MMLU 5-shot (\%)          & $45.30$ & $\mathbf{44.65}$ & $40.85$ \\
WikiText-2 PPL ($\downarrow$) & $\phantom{0}5.47$  &
$\phantom{0}\mathbf{5.58}$  & $\phantom{0}8.11$  \\
\bottomrule
\end{tabular}
} \caption{Flagged against non-flagged layer groups at matched
$\epsilon$, for the settings whose metrics do not fit the format of
Tab.~\ref{tab:main}. \textit{Flagged} is the layer group the
diagnostic selects; \textit{Non-flagged} truncates the same number of
layers elsewhere in the same model. Layer sets in braces.}
\label{tab:qa_flagged}
\end{table}

\textbf{RQ2: Where does redundancy live, and does it depend on the task?}
The answer depends on the task type. For the three classification
models (DistilBERT, ViT-B, Swin-B) and for generation on Llama-2-7B,
the diagnostic flags the early to middle layers as the most
compressible, though not the very earliest: DistilBERT at $\{0,1,2\}$,
Swin-B at $\{4,\ldots,7\}$, ViT-B at $\{6,7,8\}$, and Llama-2-7B at
$\{8,\ldots,11\}$ (Tab.~\ref{tab:main}). For dense prediction
(DPT-Large on ADE20K) the pattern shifts to the late layers
$\{16,\ldots,19\}$, where truncation preserves mIoU exactly.
Extractive QA joins dense prediction rather than classification: the
diagnostic flags RoBERTa-base at $\{8,\ldots,11\}$ of $12$ layers and
MiniLM at $\{9,10,11\}$, replicating across two architectures and two
datasets (Tab.~\ref{tab:qa_flagged}a). A single reading covers both
reversals. Dense prediction and span extraction need token-level
detail preserved through the early and middle blocks, so heads there
use more of their rank budget and the redundancy shifts late; in
sentiment and image classification and in generation the early and
middle layers do generic token-level processing and the heavy
task-specific work is concentrated in the late blocks.

\textbf{RQ3: Is the diagnostic predictive?}
Truncating the diagnostic-selected layers preserves task quality
across all seven models (Tab.~\ref{tab:main},
Tab.~\ref{tab:qa_flagged}). The clearest evidence comes from the QA
models, where the metric has the most headroom: at $\epsilon=0.3$,
truncating the flagged layers costs at most $1.75$ F1 in every
setting, while truncating the same number of layers elsewhere in the
same model costs up to $50$ F1 (MiniLM on NewsQA, $55.58$ against
$5.94$). Llama-2-7B separates the same way on probes denser than
ROUGE-L (Tab.~\ref{tab:qa_flagged}b): at the flagged layers MMLU falls
$0.65$ points and perplexity rises $0.11$, while the non-flagged
layers cost $4.45$ MMLU and $2.64$ perplexity. Effects are smaller on
the remaining models: Swin-B drops $0.03$ Top-1 at $-1.29\%$ FLOPs,
ViT-B drops $3.14$ Top-1 at $-0.76\%$, and DPT-Large is unchanged in
mIoU at $-1.54\%$. DistilBERT scores $+1.00$ accuracy above baseline
at $-0.83\%$ FLOPs, but we do not read this as an improvement:
differences of this size are small for a five-class metric and
Tab.~\ref{tab:sweep_bert} in \S\ref{app:sweep} shows several
layer choices above baseline, so we claim only that quality is
preserved. The metric does detect damage when there is damage:
sweeping $\epsilon$ from $0.1$ to $1.0$ on DistilBERT layers
$\{0,\ldots,5\}$, accuracy falls monotonically from $50.5$ to $38.2$.
Composing with WANDA at $10\%$ sparsity adds roughly nine percentage
points of FLOP reduction while changing quality by at most $0.13$ F1
on RoBERTa-base/NewsQA. Most of the composed figure is WANDA's; the
point of the composition is that the two methods remove different
weights, so their savings add rather than compete. The diagnostic also
predicts where compression \emph{fails} (\S\ref{app:sweep}),
its layer-level answer is stable under distribution shift, calibration
draw, and calibration size (\S\ref{app:robustness}), and within
the flagged layers the quality-versus-$\epsilon$ frontier is smooth
rather than a cliff (\S\ref{app:frontier}).

\subsection{Layer-Wise Sweep}
\label{app:sweep}

Tabs.~\ref{tab:sweep_vit}, \ref{tab:sweep_bert}, \ref{tab:sweep_swin},
\ref{tab:sweep_dpt} report the layer-wise sweep at $\epsilon=0.1$ for
four of the seven models. These results back the layer-selection
claims in \S\ref{sec:rq}: the configuration the diagnostic flags
as most redundant (\textbf{bold} row) gives the best accuracy-FLOPs
tradeoff, and truncating in the wrong layers degrades performance
sharply.

\begin{table}[!h]
\centering
\small
\setlength{\tabcolsep}{3pt}
\begin{tabular}{@{}lccc@{}}
\toprule
Layers & Top-1 (\%) & SVD $\Delta$ FLOPs & + WANDA \\
\midrule
$\{0,1,2\}$           & $77.22$ & $-2.47\%$ & $-11.67\%$ \\
$\{3,4,5\}$           & $77.92$ & $-1.09\%$ & $-10.43\%$ \\
$\mathbf{\{6,7,8\}}$  & $\mathbf{77.94}$ & $\mathbf{-0.76\%}$ &
$\mathbf{-10.13\%}$ \\
$\{9,10,11\}$         & $71.63$ & $-0.62\%$ & $-10.00\%$ \\
$\{0,\ldots,5\}$      & $71.07$ & $-3.56\%$ & $-12.66\%$ \\
$\{6,\ldots,11\}$     & $67.60$ & $-1.38\%$ & $-10.69\%$ \\
$\{0,\ldots,11\}$     & $47.89$ & $-4.94\%$ & $-13.90\%$ \\
\bottomrule
\end{tabular}
\caption{ViT-B layer-wise sweep on ImageNet-1K at $\epsilon=0.1$.
Truncating middle layers $\{6,7,8\}$ preserves Top-1 within 3.14
points of the baseline (81.08). Pushing truncation to layers
$\{9,10,11\}$ or wider drops accuracy sharply.}
\label{tab:sweep_vit}
\end{table}

\begin{table}[!h]
\centering
\small
\setlength{\tabcolsep}{3pt}
\begin{tabular}{@{}lccc@{}}
\toprule
Layers & Acc. (\%) & SVD $\Delta$ FLOPs & + WANDA \\
\midrule
$\{0,1\}$       & $50.63$ & $-1.79\%$ & $-11.26\%$ \\
$\{2,3\}$       & $49.64$ & $-1.53\%$ & $-11.03\%$ \\
$\mathbf{\{0,1,2\}}$     & $\mathbf{50.95}$ & $\mathbf{-0.83\%}$ &
$\mathbf{-10.39\%}$ \\
$\{3,4,5\}$     & $50.62$ & $-1.57\%$ & $-11.06\%$ \\
$\{0,\ldots,5\}$ & $50.50$ & $-4.15\%$ & $-13.38\%$ \\
\bottomrule
\end{tabular}
\caption{DistilBERT layer-wise sweep on SST-5 at $\epsilon=0.1$.
Several layer choices land slightly above the baseline (49.95),
including the flagged group $\{0,1,2\}$; we read this as quality being
preserved rather than improved.}
\label{tab:sweep_bert}
\end{table}

\begin{table}[!h]
\centering
\small
\setlength{\tabcolsep}{3pt}
\begin{tabular}{@{}lccc@{}}
\toprule
Layers & Top-1 (\%) & SVD $\Delta$ FLOPs & + WANDA \\
\midrule
$\{0,\ldots,3\}$       & $84.68$ & $-1.70\%$ & $-11.06\%$ \\
$\mathbf{\{4,\ldots,7\}}$ & $\mathbf{84.68}$ & $\mathbf{-1.29\%}$ &
$\mathbf{-10.65\%}$ \\
$\{8,\ldots,11\}$      & $84.64$ & $-0.62\%$ & $-10.05\%$ \\
$\{12,\ldots,15\}$     & $84.24$ & $-0.58\%$ & $-10.01\%$ \\
$\{16,\ldots,19\}$     & $83.41$ & $-0.50\%$ & $-9.94\%$ \\
$\{20,\ldots,23\}$     & $80.88$ & $-0.54\%$ & $-9.98\%$ \\
all 24 layers          & $76.33$ & $-5.22\%$ & $-14.26\%$ \\
\bottomrule
\end{tabular}
\caption{Swin-B layer-wise sweep on ImageNet-1K at $\epsilon=0.1$. Accuracy is well preserved in the first three blocks (layers 0-11) and degrades monotonically as truncation moves to later blocks.}
\label{tab:sweep_swin}
\end{table}

\begin{table}[!h]
\centering
\small
\setlength{\tabcolsep}{3pt}
\begin{tabular}{@{}lccc@{}}
\toprule
Layers & mIoU & SVD $\Delta$ FLOPs & + WANDA \\
\midrule
$\{0,\ldots,3\}$       & $44.29$ & $-2.00\%$ & $-8.82\%$ \\
$\{4,\ldots,7\}$       & $45.38$ & $-1.55\%$ & $-8.40\%$ \\
$\{8,\ldots,11\}$      & $45.73$ & $-1.52\%$ & $-8.37\%$ \\
$\{12,\ldots,15\}$     & $45.95$ & $-1.70\%$ & $-8.54\%$ \\
$\mathbf{\{16,\ldots,19\}}$ & $\mathbf{46.02}$ & $\mathbf{-1.54\%}$ &
$\mathbf{-8.40\%}$ \\
$\{20,\ldots,23\}$     & $45.99$ & $-1.01\%$ & $-7.91\%$ \\
all 24 layers          & $43.32$ & $-9.31\%$ & $-15.56\%$ \\
\bottomrule
\end{tabular}
\caption{DPT-Large layer-wise sweep on ADE20K at $\epsilon=0.1$. mIoU
rises monotonically as the truncation window moves later, the opposite
of ViT-B and Swin-B; late layers $\{16,\ldots,19\}$ preserve baseline
mIoU.}
\label{tab:sweep_dpt}
\end{table}

\subsection{Robustness of the Diagnostic}
\label{app:robustness}

Our pipeline is deterministic. We use public pretrained checkpoints,
we do no training or finetuning, and we run only forward passes for
calibration and evaluation, so no random seed enters the
decomposition, the mode selection, or the metric, and repeated runs
are bit-identical. The one real source of variation is which examples
land in the calibration set. We separate the three ways a calibration
set can vary and test each: the distribution it is drawn from, which
examples happen to be drawn, and how many.

\paragraph{Different distribution.}
We take RoBERTa-base finetuned on SQuAD 2.0, calibrate the diagnostic
only on SQuAD 2.0 data, and evaluate on NewsQA (baseline F1 $57.95$ in
this shifted setting). We repeat the experiment on MiniLM at
$\epsilon=0.6$, where truncation is most destructive and instability
would be easiest to see (baseline F1 $48.39$).
Tab.~\ref{tab:robust_shift} shows the layers flagged redundant stay
redundant under shift and the layers flagged critical stay critical.

\begin{table}[!h]
\centering
\small
\setlength{\tabcolsep}{3pt}
\begin{tabular}{@{}lcc@{}}
\toprule
\multicolumn{3}{@{}l@{}}{\textit{RoBERTa-base, $\epsilon=0.3$ (F1)}}
\\
\midrule
Layers truncated & calib $1\%$ & calib $15\%$ \\
\midrule
$\{8,\ldots,11\}$ \textit{(flagged redundant)} & $56.25$ & $56.03$ \\
$\{0,\ldots,3\}$                               & $51.55$ & $53.34$ \\
$\{4,\ldots,7\}$ \textit{(flagged critical)}   & $25.87$ & $25.49$ \\
\midrule
\multicolumn{3}{@{}l@{}}{\textit{MiniLM, $\epsilon=0.6$ (F1)}} \\
\midrule
$\{9,10,11\}$ \textit{(flagged redundant)}     &
\multicolumn{2}{c}{$37.45$} \\
$\{3,4,5\}$                                    &
\multicolumn{2}{c}{$15.88$} \\
$\{6,7,8\}$ \textit{(flagged critical)}        &
\multicolumn{2}{c}{$5.73$}  \\
\bottomrule
\end{tabular}
\caption{Robustness to distribution shift. The diagnostic is
calibrated on SQuAD 2.0 and evaluated on NewsQA. On MiniLM the flagged
layers hold up even when $60\%$ of the importance mass is discarded,
while truncating the critical layers $\{6,7,8\}$ costs over $42$ F1.}
\label{tab:robust_shift}
\end{table}

\paragraph{Different examples, same size.}
We rebuilt the calibration set five times with different random draws
of the same size on DistilBERT/SST-5 and reran the whole diagnostic
each time. Tab.~\ref{tab:robust_draws} reports $\epsilon=0.9$, the
most aggressive truncation we ran, because that is where instability
would be easiest to see. The spread across draws is under $0.2$
accuracy and the ranking of layer groups is identical in all five
draws. The same holds at every $\epsilon$ from $0.5$ to $0.9$, where
the largest standard deviation anywhere in the sweep is $0.39$.

\begin{table}[!h]
\centering
\small
\setlength{\tabcolsep}{3pt}
\begin{tabular}{@{}lcc@{}}
\toprule
Layers truncated & Accuracy (mean $\pm$ std) & Min-Max \\
\midrule
$\{0,1\}$ \textit{(flagged)} & $45.73 \pm 0.06$ & $45.70$-$45.84$ \\
$\{2,3\}$                    & $44.49 \pm 0.16$ & $44.21$-$44.57$ \\
$\{4,5\}$                    & $44.25 \pm 0.12$ & $44.07$-$44.39$ \\
\bottomrule
\end{tabular}
\caption{Robustness to the calibration draw. DistilBERT on SST-5 at $\epsilon=0.9$, mean over five independent draws of the same size.}
\label{tab:robust_draws}
\end{table}

\paragraph{Calibration size.}
We swept the calibration budget on RoBERTa-base/SQuAD 2.0 across more
than two orders of magnitude at $\epsilon=0.3$
(Tab.~\ref{tab:robust_size}). The flagged layers vary by only $0.25$
F1 as the calibration set grows from $100$ to $19{,}548$ examples, and
the ordering of the layer groups is identical at every size. The same
holds on MiniLM under distribution shift, where at every $\epsilon$
from $0.2$ to $0.6$ all four layer groups return the same F1 whether
we calibrate with $1$, $16$, $64$, or $256$ examples. A very small
calibration budget therefore recovers the same mode selection, which
also means the diagnostic is cheap to run.

\begin{table}[!h]
\centering
\small
\setlength{\tabcolsep}{3pt}
\begin{tabular}{@{}lccc@{}}
\toprule
Calib.\ size (share) & $\{0,\ldots,3\}$ & $\{4,\ldots,7\}$ &
$\{8,\ldots,11\}$ \\
\midrule
$100$ ($0.08\%$) & $76.87$ & $56.95$ & $\mathbf{83.10}$ \\
$1{,}303$ ($1\%$) & $77.06$ & $52.23$ & $\mathbf{82.85}$ \\
$6{,}516$ ($5\%$) & $75.23$ & $51.76$ & $\mathbf{82.86}$ \\
$13{,}032$ ($10\%$) & $75.41$ & $53.51$ & $\mathbf{82.97}$ \\
$19{,}548$ ($15\%$) & $75.58$ & $53.57$ & $\mathbf{82.95}$ \\
\bottomrule
\end{tabular}
\caption{Robustness to calibration size. RoBERTa-base on SQuAD 2.0 at
$\epsilon=0.3$ (F1, baseline $83.01$). Bold column is the layer group
the diagnostic flags as redundant.}
\label{tab:robust_size}
\end{table}

\subsection{Rank-Energy Frontier in the Flagged Layers}
\label{app:frontier}

The energy threshold $\epsilon$ controls how much of the importance
mass is discarded, and therefore how many modes each head keeps.
Tab.~\ref{tab:epsilon_frontier} sweeps it on the layers the diagnostic
flags in RoBERTa-base/SQuAD 2.0. The mean retained rank falls from
$49$ to $38$ to $29$ of $64$ as $\epsilon$ goes from $0.1$ to $0.3$,
while F1 moves from $83.24$ to $82.85$ against a baseline of $83.01$;
discarding more than half of the average head's rank budget therefore
costs $0.16$ F1.

Two things follow. First, the redundancy the diagnostic reports is
large in absolute terms rather than a marginal trim at the tail of the
spectrum, which is what the small whole-model FLOP percentages in
Tab.~\ref{tab:main} might otherwise suggest. Second, $\epsilon$
behaves as a usable dial rather than a threshold that has to be tuned
precisely: within the flagged layers quality degrades gradually, so it
is the layer selection rather than the exact energy level that
determines whether truncation is safe. The contrast with
Tab.~\ref{tab:qa_flagged}, where the same $\epsilon$ applied to
non-flagged layers costs up to $50$ F1, is the point.

\begin{table}[!h]
\centering
\small
\setlength{\tabcolsep}{3pt}
\begin{tabular}{@{}lccc@{}}
\toprule
 & $\epsilon=0.1$ & $\epsilon=0.2$ & $\epsilon=0.3$ \\
\midrule
F1                          & $83.24$ & $83.09$ & $82.85$ \\
Mean retained rank $r_h$    & $49$    & $38$    & $29$    \\
\bottomrule
\end{tabular}
\caption{RoBERTa-base on SQuAD 2.0, flagged layers $\{8,\ldots,11\}$
(baseline F1 $83.01$, $d_h=64$). The quality-versus-$\epsilon$
frontier is smooth: more than half of the average head's rank budget
is discarded at $\epsilon=0.3$ for a loss of $0.16$ F1.}
\label{tab:epsilon_frontier}
\end{table}

\section{Conclusion}
\label{sec:conclusion}

We presented SVD-Attention, a data-aware spectral diagnostic that
scores each singular mode of the per-head interaction matrix $M_h$ by
its singular value and the geometric mean of query- and key-side
activation variances, and discards the rest without finetuning. Two
findings emerge. First, the layer at which redundancy concentrates
depends on task type: early to middle for sentiment and image
classification and for generation, late for dense prediction and
extractive question answering. Second, acting on the diagnostic yields
measurable FLOP reductions across all seven models we tested and
composes with weight-level pruning such as WANDA. The diagnostic also
predicts where compression fails, positioning SVD-Attention as a
measurement tool first and a compression tool second.

\section{Limitations}
\label{sec:limitations}

We acknowledge several limitations. First, our compression gains are
modest in absolute terms compared with token-level methods, primarily
because we compress only multi-head self-attention. Our contribution
is the diagnostic and its no-finetuning property, not a
state-of-the-art compression ratio. Second, we report analytic FLOPs.
Turning them into wall-clock speedup requires an attention kernel at
the reduced per-head dimension, since padding the discarded dimensions
back erases the saving. Third, under RoPE the query-key interaction is
a relative-position family rather than a single matrix. We argue in
\S\ref{app:rope} that mode selection is nonetheless
position-independent and that the truncation error is bounded
uniformly in position, but we do not claim an exact factorization of
the full family and we do not treat grouped-query attention. Finally,
we do not evaluate interactions with quantization or with
attention-pattern modifications such as linear attention; we expect
the gains to compose, but this has not been verified empirically.



\section*{Acknowledgments}
This work was conducted in part during an internship at Dolby
Laboratories. This work was also supported in part by CoCoSys, one of
seven centers in JUMP 2.0, a Semiconductor Research Corporation (SRC)
program sponsored by DARPA.

\bibliography{custom}

\end{document}